\documentclass[letterpaper, 10 pt, journal, twoside]{ieeetran}

\usepackage{graphics} 
\usepackage{epsfig} 
\usepackage{times} 
\usepackage{amsmath} 
\usepackage{amssymb}  
\usepackage[bottom]{footmisc}

\makeatletter
\let\MYcaption\@makecaption
\makeatother

\usepackage[font=footnotesize,aboveskip=12pt]{subcaption}

\makeatletter
\let\@makecaption\MYcaption
\makeatother

\makeatletter
\let\NAT@parse\undefined
\makeatother

\usepackage[noadjust]{cite}

\usepackage{hyperref}
\usepackage{cleveref}
\Crefname{equation}{Eq.}{Eqs.}
\Crefname{figure}{Fig.}{Figs.}
\Crefname{tabular}{Tab.}{Tabs.}
\crefname{appsec}{Appendix}{Appendices}

\title{Evolved Neuromorphic Control for High Speed Divergence-based Landings of MAVs}

\author{Jesse J. Hagenaars$^1$, Federico Paredes-Vall\'es$^1$, Sander M. Boht\'e$^2$, and Guido C.\thinspace H.\thinspace E. de Croon$^1$
\thanks{Manuscript received: February 23, 2020; Revised May 19, 2020; Accepted July 19, 2020.}%
\thanks{This paper was recommended for publication by Editor Jonathan Roberts upon evaluation of the Associate Editor and Reviewers' comments.}%
\thanks{$^1$J.\thinspace J. Hagenaars, F. Paredes-Vall\'es, and G.\thinspace C.\thinspace H.\thinspace E. de Croon are with the Micro Air Vehicle Laboratory, Faculty of Aerospace Engineering, Delft University of Technology, The Netherlands (\href{mailto:j.j.hagenaars@tudelft.nl}{\texttt{j.j.hagenaars@tudelft.nl}}).}%
\thanks{$^2$S.\thinspace M. Boht\'e is with the Machine Learning Group, Centrum Wiskunde $\&$ Informatica, The Netherlands.}%
\thanks{Digital Object Identifier (DOI): see top of this page.}%
}

\begin{document}

\maketitle

\begin{abstract}
Flying insects are capable of vision-based navigation in cluttered environments, reliably avoiding obstacles through fast and agile maneuvers, while being very efficient in the processing of visual stimuli. Meanwhile, autonomous micro air vehicles still lag far behind their biological counterparts, displaying inferior performance at a much higher energy consumption. In light of this, we want to mimic flying insects in terms of their processing capabilities, and consequently show the efficiency of this approach in the real world. This letter does so through evolving spiking neural networks for controlling landings of micro air vehicles using optical flow divergence from a downward-looking camera. We demonstrate that the resulting neuromorphic controllers transfer robustly from a highly abstracted simulation to the real world, performing fast and safe landings while keeping network spike rate minimal. Furthermore, we provide insight into the resources required for successfully solving the problem of divergence-based landing, showing that high-resolution control can be learned with only a single spiking neuron. To the best of our knowledge, this work is the first to integrate spiking neural networks in the control loop of a real-world flying robot. Videos of the experiments can be found at \url{https://bit.ly/neuro-controller}.
\end{abstract}


\section{Introduction}

\IEEEPARstart{F}{lying} insects are everything we would like micro air vehicles (MAVs) to be: units that can navigate autonomously in cluttered environments through fast and agile maneuvers, despite being strongly limited in computational and energy resources. Like most animals that can see, these insects rely heavily on patterns of visual motion, or \textit{optical flow}~\cite{gibson_perception_1950}, for many important behaviors. During landing, for instance, honeybees maintain a constant rate of expansion, or \textit{divergence}, of the optical flow field to ensure a smooth approach~\cite{baird_universal_2013}.

Insects perceive visual motion in a \textit{spike-based} manner through light-sensitive cells and networks of interconnected neurons that react to brightness changes in the environment~\cite{posch_retinomorphic_2014}. The sparsity and asynchronicity of such a spike-driven approach have inspired researchers to come up with artificial substitutes, referred to as \textit{neuromorphic}, that could potentially be used by insect-scale MAVs~\cite{de_croon_design_2009,ma_controlled_2013} for efficient vision-based navigation. Event cameras~\cite{gallego_event-based_2019}, whose pixels register brightness changes as events, take the place of the retina. \textit{Spiking neural networks} (SNNs)~\cite{maass_networks_1997} assume the role of the underlying networks, subsequently transforming these event streams into estimates of visual motion.

Although the interest for event cameras is growing rapidly in the field of robotics~\cite{gallego_event-based_2019}, SNNs have not yet become widespread in control applications. The cause of this lies partially in the difficulty of training: the discrete spiking nature of SNNs severely limits the use of gradient-based optimization algorithms. Instead, most learning is based on the relative timing of spikes, often in combination with a surrogate gradient~\cite{tavanaei_deep_2019} or global reward signal~\cite{fremaux_neuromodulated_2016} to allow the specification of desired behavior or goals. As far as robot control is concerned, these learning rules currently seem to be limited to either simulated applications~\cite{clawson_spiking_2016,bing_indirect_2020} or simple real-world problems resembling classification~\cite{zhao_brain-inspired_2018}.

\textit{Artificial neural networks} (ANNs), on the other hand, have been employed successfully for real-world vision-based control. For instance,~\cite{scheper_evolution_2020} used \textit{neuroevolution}~\cite{floreano_neuroevolution:_2008} to optimize ANNs for performing divergence-based landings of MAVs. This work aims to demonstrate that we can evolve SNNs to solve the same control problem while keeping energy consumption at a minimum. The generality of evolutionary algorithms with respect to the characteristics of the evolved individuals~\cite{fogel_advantages_1997} makes this, in our opinion, the most promising current approach to SNN learning.

\begin{figure*}[!t]
    \centering
    \includegraphics[width=0.79\textwidth]{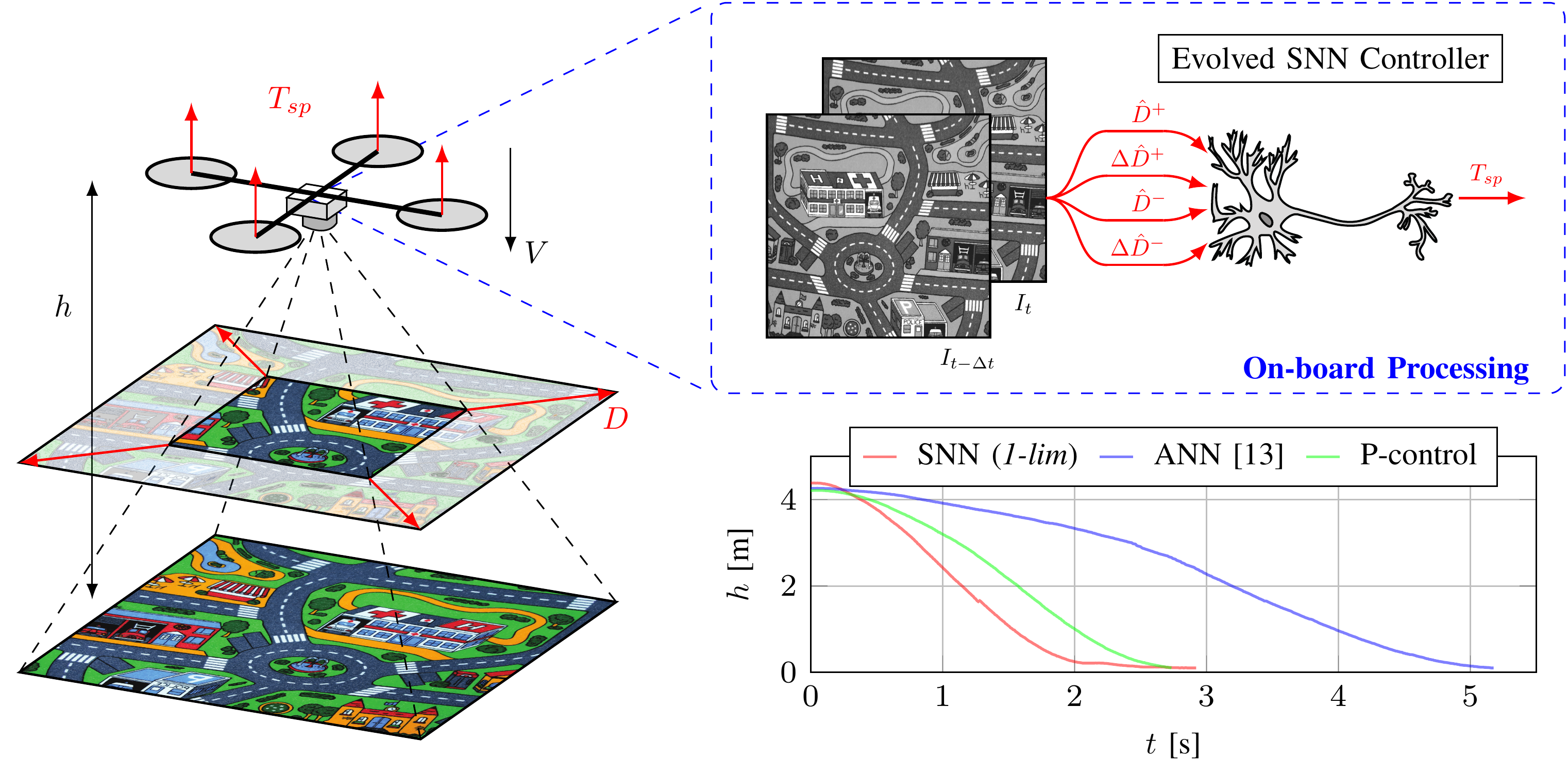}
    \caption{Overview of the proposed system. An MAV with downward-facing camera is controlled to perform vertical landings based on the divergence $D$ of the optical flow field. As the MAV moves towards the surface, its field-of-view covers a smaller portion of the original pattern, and distances between any two tracked points on the camera's pixel array increase. This increase is proportional to $D$. Two subsequent video frames $I_{t-\Delta t}$ and $I_t$ can thus provide estimates of divergence $\hat{D}$ and its temporal derivative $\Delta\hat{D}$, which can subsequently be used by an evolved SNN controller to regulate the thrust setpoint $T_{sp}$. Our controller compares favorably against a state-of-the-art ANN controllers~\cite{scheper_evolution_2020} and a proportional controller during real-world tests.}
    \label{fig:opticalflow}
\end{figure*}

This letter contains two main contributions. First, we demonstrate learned neuromorphic control for a real-world problem through evolving SNNs for performing divergence-based landings of an MAV. To the best of our knowledge, this work is the first to integrate SNNs in the control loop of a real-world flying robot. Second, we study how to substantially reduce the spike rate of the SNN controller, corresponding to considerable energy savings if it were to be run on neuromorphic hardware. Besides investigating the effect of pruning neurons (as also done in~\cite{iglesias_dynamics_2005,dora_two_2015}), we introduce the inclusion of network spike rate as an objective in the multi-objective neuroevolution. \Cref{fig:opticalflow} presents an overview of the proposed system.

The remainder of this letter is structured as follows. \Cref{sec:related} provides related work concerning robot learning. The control problem, SNN configuration, and learning procedure are discussed in \Cref{sec:meth}. Next, \Cref{sec:exp} covers the performed experiments and lists their findings. Conclusions drawn from these findings are then stated in \Cref{sec:conc}.

\section{Related Work}
\label{sec:related}

Not all approaches are equally well suited to the problem of learning real-world robot control, which can be characterized as the optimization of some behavioral function in a complex, uncertain environment. One of the more popular paradigms for solving these kinds of problems has been \textit{reinforcement learning} (RL) in combination with deep ANNs~\cite{hwangbo_control_2017}. In pursuit of more efficient methods, we investigate SNNs as an alternative to these deep ANNs, knowing that the respective energy savings can be as large as an order of magnitude for comparable networks~\cite{pfeiffer_deep_2018}. Through the related work, we aim to show the immaturity of current RL-inspired approaches to SNN learning, called \textit{reward-modulated}, as well as the promise of neuroevolution.

\subsection{Reinforcement Learning in SNNs}

SNNs trained through reward-modulated learning have so far only been successfully applied to problems that are either relatively simple or simulated. For instance, \cite{zhao_brain-inspired_2018} succeeds in training an SNN for a real-world MAV obstacle avoidance task using a reward-modulated rule, but only after the problem has been preprocessed to a much simpler (almost one-to-one) mapping between discrete inputs and outputs.

In simulation,~\cite{bing_indirect_2020} demonstrates vision-based neuromorphic lane-keeping control of a two-wheeled robot. Although a reward-modulated rule is used for learning, the task is set up in such a way that its complexity remains limited: rewards are tailored to each individual neuron, so that increased firing inevitably results in a self-centering policy. In~\cite{clawson_spiking_2016}, the authors employ the same learning rule for training an SNN to control a simulated robotic insect. Reward is based on the deviation from an externally generated trajectory, however, making it essentially a lane-keeping task.

\subsection{Neuroevolution for Robot Control}

Reviews of the field of neuroevolution show its promise for learning in ANNs~\cite{stanley_designing_2019} and SNNs~\cite{floreano_neuroevolution:_2008}. Furthermore, neuroevolution exhibits qualities relevant to real-world learning: it scales well in terms of parameter space and compute, and can even be more sample efficient than RL~\cite{stanley_designing_2019}. 
So far, evolved SNN controllers have only been successfully applied to simulated MAVs~\cite{howard_evolving_2014,howard_towards_2016}, or rudimentary real-world ground robots~\cite{bing_survey_2018}. ANN neuroevolution, on the other hand, has been applied to more complex real-world problems. For example, the authors of \cite{scheper_evolution_2020} evolve ANNs to learn event-based optical flow control of a real-world, landing MAV. It is shown that a small, three-layer network is sufficient to perform high-resolution control, with only the weights being evolved. Following their success, we directly extend this approach to SNNs, in order to obtain a more energy-efficient solution.

\section{Methodology}
\label{sec:meth}

\subsection{Divergence of the Optical Flow Field}

In this work, we use the optical flow formulation from \cite{de_croon_optic-flow_2013}, which assumes a downward-looking camera over a static planar surface, as depicted in \Cref{fig:opticalflow}. With this configuration, moving the camera along the $Z$-axis (as for a vertical MAV landing) causes an optical flow, in this case divergence, to be perceived. Physically, this divergence corresponds to the ratio of vertical velocity and height above the surface, or $D = V/h$. To estimate divergence from a camera, we can use the relative, temporal variation in the distance between tracked image points (corners)~\cite{ho_adaptive_2018}. Referred to as \textit{size divergence}, this method results in a computationally efficient and reliable estimate of divergence $\hat{D}$, when averaged over a set of $N_{D}$ pairs of points:
\begin{equation}
\label{eq:of6}
    \hat{D}(t) = \frac{1}{N_D}\sum_{i = 1}^{N_D} \frac{1}{\Delta t}\frac{l_i(t - \Delta t) - l_i(t)}{l_i(t - \Delta t)}
\end{equation}

\noindent{with $\Delta t$ the time step, and $l_i(t)$ the distance between a pair of tracked points at time $t$. The proposed SNN controller receives as input an estimate of divergence $\hat{D}$, as well as its temporal derivative $\Delta\hat{D}$.}

\subsection{Spiking Neural Network Architecture}

In SNNs, neurons are connected through \textit{synapses}, which have a certain weight. Incoming spikes contribute to the \textit{membrane potential} $u_i(t)$ of a neuron in an additive or subtractive manner. In case no inputs are received, $u_i(t)$ decays to a resting potential $u_\mathit{rest}$. On the other hand, if the quantity of inputs is large enough to push the potential above a threshold $\theta_i$, the neuron itself emits a spike  $s_i$, after which the potential is reset to $u_\mathit{rest}$.

The neuron model employed in this work is the often-used \textit{leaky integrate-and-fire} (LIF)~\cite{stein_theoretical_1965}. Discretizing this model using forward Euler leaves us with the following equation for the membrane potential:
\begin{equation}
\label{eq:lif}
    u_i(t) = u_i(t - \Delta t) \cdot \tau_{u_i} + \alpha_{u_i} i_i(t)
\end{equation}

\noindent{where we assumed $u_\mathit{rest}=0$, and take the membrane decay as a factor $\tau_{u_i}$. $i_i(t)$ is the forcing function working on neuron $i$, which corresponds to the incoming spikes multiplied by their respective synaptic weights, i.e., $i_i(t) = \sum_j w_{ij}s_j(t)$, or to incoming currents $c_j(t)$, i.e., $i_i(t) = \sum_j w_{ij}c_j(t)$. The influence of the forcing function on the membrane potential is scaled with a constant $\alpha_{u_i}$.}

To prevent excessive firing while ensuring responsiveness to small/low-frequency inputs, $\theta_i$ can be made dependent on the neuron's firing rate, resulting in an adaptive LIF~\cite{liu_spike-frequency_2001}:
\begin{equation}
\label{eq:thresh}
    \theta_i(t) = \theta_i(t - \Delta t) \cdot \tau_{\theta_i} + \alpha_{\theta_i} s_i(t)
\end{equation}

\noindent{with $\tau_{\theta_i}$ being the corresponding decay factor, and $\alpha_{\theta_i}$ the constant scaling the emitted spike.}

The binary nature of SNNs requires functions that transform real-valued signals to binary spikes and vice-versa, i.e., \textit{encodings} and \textit{decodings}. This work makes use of a pair of non-spiking neurons per input observation, one for positive and one for negative values, with at most one of the two neurons active at any given time. The proportional currents $c^+_i(t)$ and $c^-_i(t)$ coming out of the respective neurons can be expressed as:
\begin{equation}
\begin{split}
\label{eq:enc1}
    c^+_i(t) &= \lvert \max(0, o_i(t)) \rvert \\
    c^-_i(t) &= \lvert \min(0, o_i(t)) \rvert
\end{split}
\end{equation}

\noindent{with $o_i(t)$ the observation variable belonging to neuron $i$.}

For decoding binary spikes to real-valued scalars $a_i(t)$ (actions) in a range $[r_1, r_2]$, the \textit{spike trace} $X_i(t)$, which is essentially a low-pass filter over a neuron's emitted spikes, can be combined with a simple scaling:
\begin{equation}
\begin{split}
\label{eq:trace}
    a_i(t) &= r_1 + (r_2 - r_1) \cdot X_i(t) \\
    X_i(t) &= X_i(t - \Delta t) \cdot \tau_{x_i} + \alpha_{x_i} s_i(t)
\end{split}
\end{equation}

The SNN used for the control task in this work is kept relatively simple, with only a single hidden layer of not more than 20 adaptive LIF neurons, and a single output LIF neuron. We consider vertical control to be one-dimensional, with the SNN controller setting the thrust. Two pairs of non-spiking neurons encode the inputs $\hat{D}$ and $\Delta\hat{D}$, as in \Cref{eq:enc1}. See \Cref{fig:opticalflow} for an illustration.

\subsection{Evolving Energy-efficient Neuromorphic Controllers}

Each evolution starts off with a randomly initialized \textit{population} of $\mu$ SNN individuals. We opt for a \textit{mutation-only} approach, given that crossover tends to work best when natural building blocks are available, and could lead to difficulties like the permutation problem when applied to neural networks~\cite{yao_evolving_1999}. Weights and hyperparameters are mutated with $P_\mathit{mut}=0.3$ according to the distributions in \Cref{tab:mutations}. \textit{Offspring} $\lambda$ is combined with the previous population and evaluated in a highly stochastic simulation environment (see \Cref{sec:env}), where the repeated evaluation (along with resampling of the environment) of the previous generation decreases the chance individuals live on only because they received `easy' environmental conditions (little noise, small delay, fast-responding motors, etc.). The \textit{fitness} of an individual consists of four objectives: time to land ($f_1$), final height ($f_2$), final vertical velocity ($f_3$), and total spike rate of the network ($f_4$). Selection is carried out using the multi-objective genetic algorithm \textit{NSGA-II}~\cite{deb_fast_2002}.

During evolution, a \textit{hall of fame} is maintained, which holds the pan-generational Pareto front (all non-dominated individuals that have ever lived). This prevents the discard of well-performing individuals across generations. After $N_\mathit{gen}$ generations, the individuals in the hall of fame are evaluated by letting them perform 250 landings in a randomized environment and quantifying the median and inter-quartile range for each evolutionary objective, giving us an idea of their robustness. The best-performing individuals are then selected for further real-world tests.

\renewcommand{\arraystretch}{1.2}
\begin{table}[tb]
    \centering
    \caption{Sampling distributions of mutated parameters}
    \label{tab:mutations}
    \begin{tabular}{c|c}
        \hline\hline
        \textbf{Parameter} & \textbf{Distribution} \\ \hline
        $w_{ij}$ & $\mathcal{U}(-w_{ij} - 0.05, 2w_{ij} + 0.05)$ \\
        $\alpha_{u_i}$, $\alpha_{\theta_i}$, $\alpha_{x_i}$ & $\mathcal{U}(\alpha_* - 2/3, \alpha_* + 2/3)$, clamped to $[0, 2]$ \\
        $\tau_{u_i}$, $\tau_{\theta_i}$, $\tau_{x_i}$ & $\mathcal{U}(\tau_* - 1/3, \tau_* + 1/3)$, clamped to $[0, 1]$ \\
        $\theta_i$ & $\mathcal{U}(\theta_{i} - 1/3, \theta_{i} + 1/3)$, clamped to $[0, 1]$ \\ \hline\hline
    \end{tabular}
\end{table}

\subsection{Randomized Vertical Simulation Environment}
\label{sec:env}

The vertical simulation environment in which individuals are evaluated makes use of domain randomization and artificial noise to improve transferability to the real world. The available observations are the divergence $\hat{D}$ and its temporal derivative $\Delta \hat{D}$. Similar to \cite{scheper_evolution_2020}, the simulated MAV is considered as a unit mass under the influence of gravity, and control happens in one dimension with the SNN controller selecting a thrust setpoint $T_{sp}$. This leads to the following dynamics model:
\begin{equation}
\label{eq:dynamics}
    \begin{split}
        h(t) &= h(t - \Delta t) + \Delta t \cdot v(t - \Delta) \\
        v(t) &= v(t - \Delta t) + \Delta t \cdot T(t - \Delta t) + W(t) \\
        T(t) &= T(t - \Delta t) + \Delta t \cdot \frac{T_{sp}\cdot g - T(t - \Delta t)}{\Delta t + \tau_T}
    \end{split}
\end{equation}

\noindent{where the altitude $h$, vertical velocity $v$, and thrust $T$ are updated using the forward Euler method, and $\tau_T$ represents the spin-up and spin-down time of the rotors. The thrust setpoint $T_{sp}$ selected by the SNN is clamped to a realistic range of acceleration for the MAV, namely $[-0.8, 0.5]$~g. Lastly, $W$ denotes vertical wind, and is given by:}
\begin{equation}
\label{eq:wind}
    W(t) = W(t - \Delta t) + \Delta t \cdot \frac{\mathcal{N}(0, \sigma^2_{\mathit{W}}) - W(t - \Delta t)}{\Delta t + \sigma_\mathit{W}}
\end{equation}

\noindent{with $\sigma_\mathit{W}=0.1$~ms$^{-1}$ being the standard deviation of the normally distributed wind.}

Noise is added to the divergence estimation according to the model in~\cite{ho_characterization_2016}. The observed divergence $\hat{D}$ is the result of adding a delay $\delta_\mathit{D}$ to the ground-truth divergence, along with white noise and proportional white noise:
\begin{equation}
\label{eq:div}
    \begin{split}
        \hat{D}(t) &= D(t - \delta_D \cdot \Delta t) + \mathcal{N}(0, \sigma^2_D) \\ &+ D(t - \delta_D \cdot \Delta t) \cdot \mathcal{N}(0, \sigma^2_{D_\mathit{prop}})
    \end{split}
\end{equation}

\noindent{where $\sigma_D$ and $\sigma_{D_\mathit{prop}}$ are the standard deviations for the added noise and proportional noise, respectively. Additionally, computational jitter is introduced in order to simulate the case in which the estimated divergence is not updated due to, for instance, insufficient corner points. Each time step, there is the probability $P_\mathit{jitter}$ that the estimated divergence from the previous step is used (for a maximum of one step).}

\renewcommand{\arraystretch}{1.2}
\begin{table}[tb]
    \centering
    \caption{Sampling distributions of environment parameters}
    \label{tab:env}
    \begin{tabular}{c|c}
        \hline\hline
        \textbf{Parameter} & \textbf{Distribution} \\ \hline
        $\delta_D$ & $\mathcal{U}(1, 4)$ steps \\
        $\sigma_D$ & $\mathcal{U}(0.05, 0.15)$ s$^{-1}$ \\
        $\sigma_{D_\mathit{prop}}$ & $\mathcal{U}(0.0, 0.25)$ s$^{-1}$ \\
        $\tau_T$ & $\mathcal{U}(0.005, 0.04)$ s \\
        $\Delta t$ & $\mathcal{U}(0.02, 0.0333)$ s \\
        $P_\mathit{jitter}$ & $\mathcal{U}(0.0, 0.2)$ \\ \hline\hline
    \end{tabular}
\end{table}

The evaluation of an individual consists of four landings, from initial altitudes $h_0 = {2, 4, 6, 8}$~m. The environment is bounded in altitude and time: $[0.05, h_0 + 5]$~m and $30$~s. Individuals start out without initial velocity and acceleration, and are left to settle for $0.5$~s. Each landing has its own, differently randomized environment, with parameters (\Cref{tab:env}) being redrawn at the start of each generation, such that all individuals experience the same four environments. Fitness is averaged across the four landings, with extra punishment for individuals that do not manage to land.

\section{Experiments}
\label{sec:exp}

\subsection{Experimental Setup}

\subsubsection{Simulation}

Per configuration, four randomly initialized populations of 100 individuals are evolved for 400 generations, after which their final halls of fame are combined. Initial synaptic weights are drawn from $\mathcal{U}(0, 1)$, and other hyperparameters are initialized as constants: $(\alpha_{u_i}, \alpha_{\theta_i}, \alpha_{x_i}) = (0.2, 0.2, 1.0)$, $(\tau_{u_i}, \tau_{\theta_i}, \tau_{x_i}) = (0.8, 0.8, 0.8)$, and $\theta_i = 0.2$.

For simulating SNNs, we used Python and the open-source PySNN\footnote{Available at \url{https://github.com/BasBuller/PySNN}} library recently developed in our lab; for performing the evolutions, we used the DEAP~\cite{fortin_deap:_2012} framework. The code for running the experiments\footnote{Available at \url{https://github.com/Huizerd/evolutionary}} and the simulation environment\footnote{Available at \url{https://github.com/Huizerd/gym-quad}} is also publicly available.

\subsubsection{Real World}

\begin{figure}[!b]
    \centering
    \includegraphics[width=0.4\textwidth]{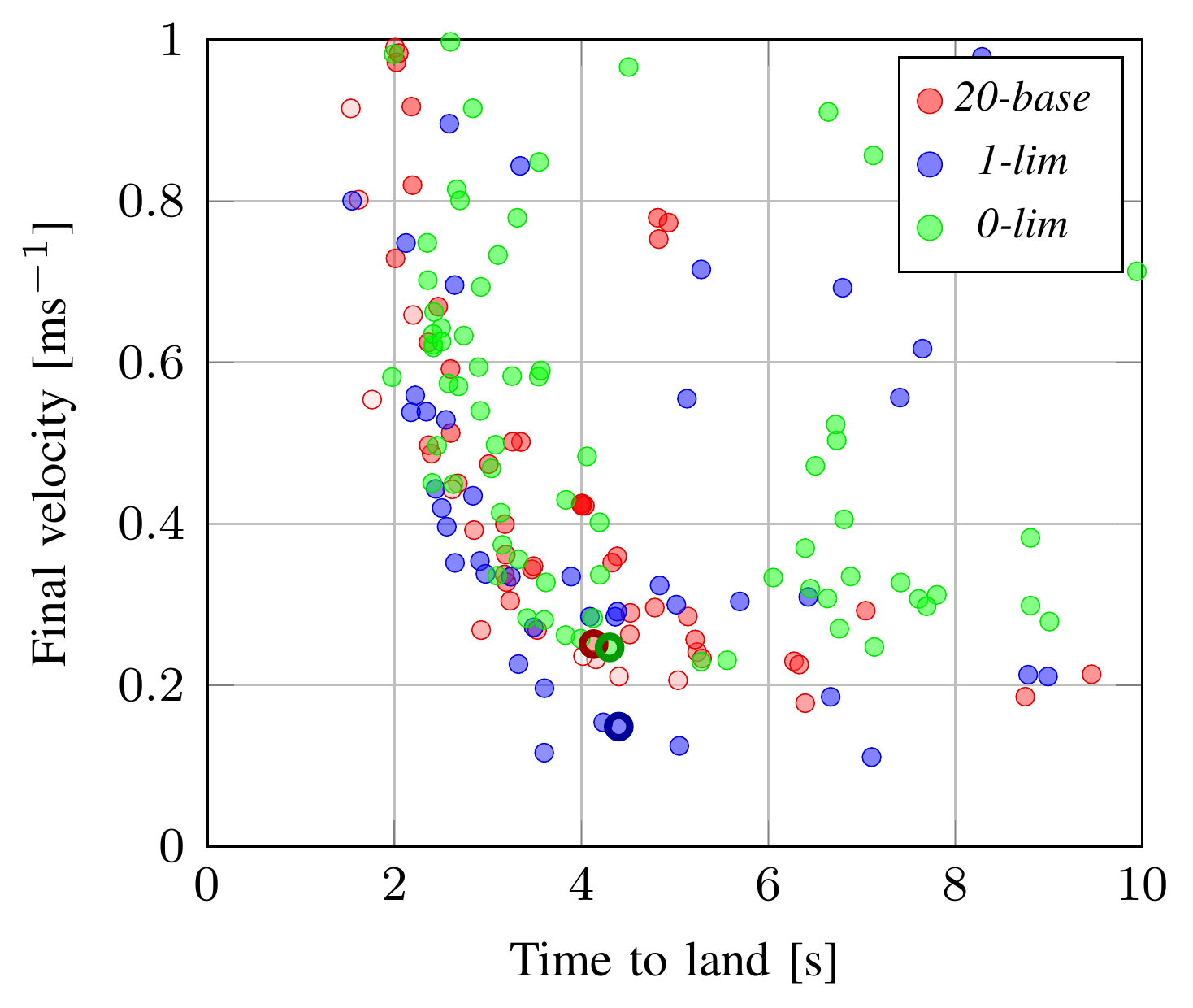}
    \caption{Pareto front (based on median performance over~250 evaluations) of individuals in the final hall of fame. The dot's color shade is proportional to the spike rate median: lighter means a higher rate. Selected individuals are indicated in bold.}
    \label{fig:pareto}
\end{figure}

The MAV used in this work is a Parrot Bebop~2 running the open-source Paparazzi\footnote{Available at \url{https://github.com/paparazzi/paparazzi}} autopilot on its $780$~MHz dual-core ARM Cortex A9 processor. To also run the SNN on board, we developed TinySNN\footnote{Available at \url{https://github.com/Huizerd/tinysnn}}: a framework for building small spiking networks in the C programming language. Its similarities with PySNN allow an almost seamless transfer of networks from simulation to the real-world hardware.

\begin{figure*}[!t]
    \centering
    \begin{subfigure}[b]{0.485\textwidth}
        \includegraphics[width=0.95\textwidth]{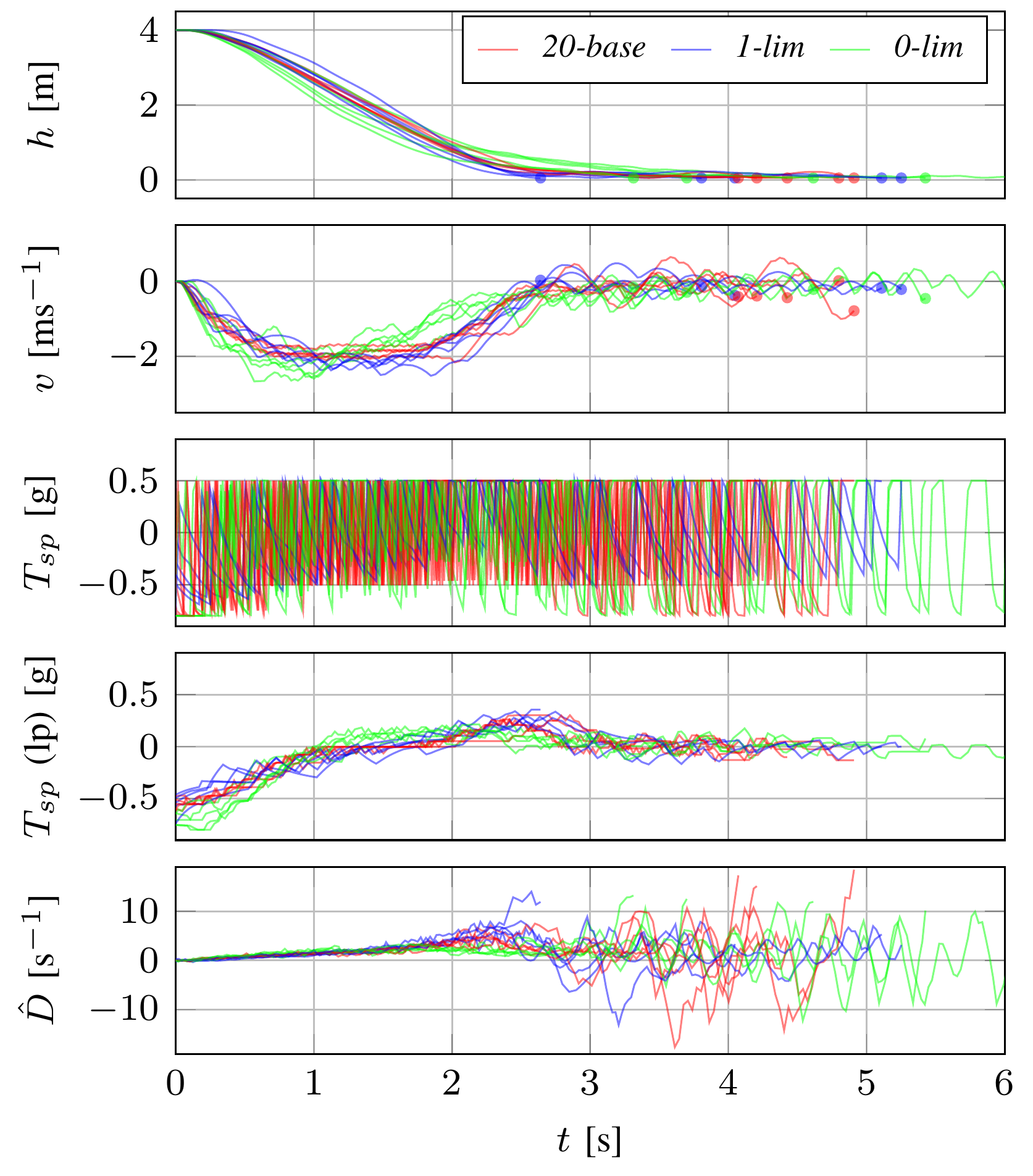}
        \vspace{-5pt}
        \caption{Five simulated runs in a randomized environment.}
        \label{fig:simperf}
    \end{subfigure}
    \hfill
    \begin{subfigure}[b]{0.485\textwidth}
        \includegraphics[width=0.95\textwidth]{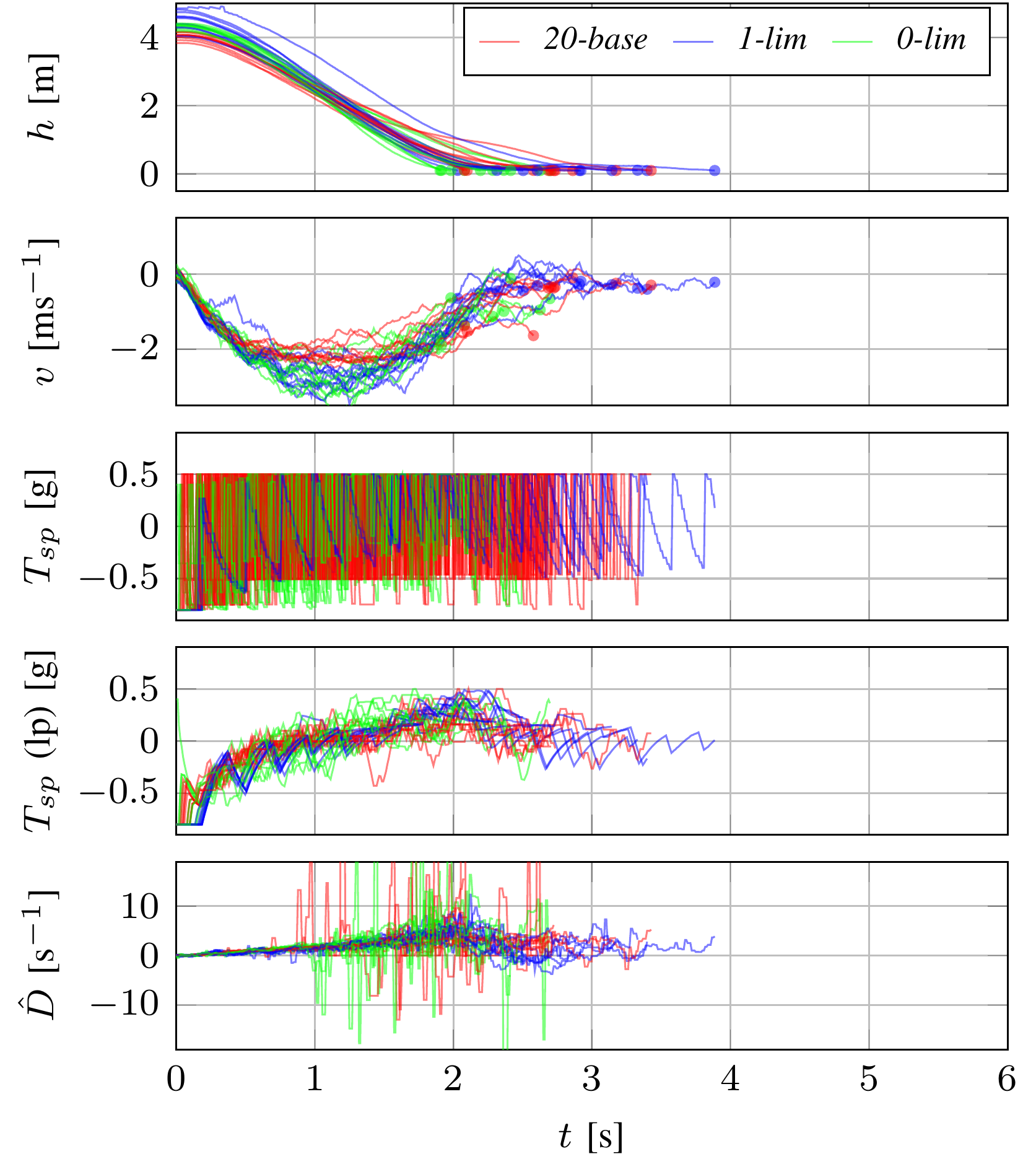}
        \vspace{-5pt}
        \caption{Ten real-world flight tests.}
        \label{fig:realperf}
    \end{subfigure}
    \caption{Height, velocity, thrust setpoint (raw and 20-step moving average) and estimated divergence for simulated and real-world landings of selected individuals. Dots in the $h$ and $v$ plot mark the end of runs.}
    \label{fig:landings}
\end{figure*}

Landings start from an initial altitude of roughly $4$~m and are ended at $0.1$~m above ground (to prevent infinite $D$ and compensate for the offset created by the MAV's landing legs at initialization). Horizontal guidance is provided by a motion capture system. Similar to~\cite{scheper_evolution_2020,pijnacker_hordijk_vertical_2018}, divergence is estimated as size divergence using the Bebop's downward-looking CMOS camera, and a FAST corner detector~\cite{rosten_machine_2006} in combination with a pyramidal Lucas-Kanade feature tracker~\cite{bouguet_pyramidal_2000}. To limit computational expense, $N_D$ is capped at 100~points. Divergence measurements are updated at a rate of approximately $45$~Hz, while the control loop implementing the divergence-based landing runs at roughly $512$~Hz.

Linearly transforming the thrust setpoint $T_{sp}$ to rotor commands leads to poor tracking performance due to unmodeled drag and nonlinear aerodynamic effects that result from a descent through the propeller downwash. To close this reality gap, a PI controller (with gains $P=0.7$ and $I=0.3$) was used to convert the thrust setpoint to motor commands~\cite{scheper_evolution_2020}.

\subsection{20 Hidden Neurons}

The first SNN configuration considered here is \textit{20-base}, which has 20 adaptive LIF neurons as hidden layer. The spike-minimizing neuroevolution allows us to start off with more neurons than necessary (\cite{scheper_evolution_2020} used eight), as redundant ones will ultimately be silenced. \Cref{fig:pareto} displays the Pareto front of evolved individuals for this configuration in red. From this front, a single individual, indicated by a bold circle, is selected for further testing. Note that objective $f_2$ (final altitude) is not shown in \Cref{fig:pareto}, as it was almost consistently minimized for all individuals.

Looking at the simulated landings performed by this individual in \Cref{fig:simperf}, we see that most landings are quite smooth (low touchdown velocity). Plots of the low-passed thrust setpoint $T_{sp}$ likewise display a small bump that suggests braking before touchdown. The raw $T_{sp}$ data, however, shows large-magnitude, high-frequency oscillations. This behavior is caused by the values of $\alpha_{x_i}$ and $\tau_{x_i}$ of the decoding, which cause instantaneous jumps and decays to maximum and (almost) minimum acceleration, respectively. Controllers that show this kind of \textit{bang-bang} behavior are unlikely to transfer well from simulation to the real world due to their dependency on motor dynamics~\cite{scheper_evolution_2020}.

When taking the selected controller to the real world, we can conclude from \Cref{fig:realperf} that this is indeed so, with higher touchdown velocities and quicker landings ($2$-$3$~s in reality versus $5$-$6$~s in simulation). The quick oscillations in $T_{sp}$ cannot be followed by the motors, leading to lower values of acceleration than actually desired. Currently, the evolutionary process has little way of accounting for this discrepancy, because the bang-bang control leads to good landings in simulation. To account for this, we constrain the mutation of $\alpha$'s (mutation magnitude halved, clamped to $[0, 1]$) and $\tau$'s (clamped to $[0.3, 1]$) in the next section.

During the simulated landings of \Cref{fig:simperf}, the spiking activity of each neuron was recorded. \Cref{fig:networks} gives the average spike rate per neuron, as well as the sign and magnitude of the connections. Looking at \textit{20-base}, the number of inactive hidden neurons and weak connections suggests it can be made much smaller. In fact, the single yellow path from input to output layer, together with the single effectively active neuron in the hidden layer, leads us to believe divergence-based landings can be performed with only a single spiking hidden neuron (\textit{1-lim}), or possibly none at all (\textit{0-lim}).

\subsection{One or No Hidden Neuron}

\Cref{fig:pareto} lets us compare the Pareto fronts for \textit{20-base}, \textit{1-lim} and \textit{0-lim}, with the latter two limited in the mutation of $\alpha$'s and $\tau$'s. The front of \textit{1-lim} outperforms both \textit{20-base} and \textit{0-lim}, suggesting there is a benefit to a significantly reduced parameter space as well as a hidden layer.

The comparison of simulated landings in \Cref{fig:simperf} shows that, even though all selected controllers perform roughly the same in terms of time to land, \textit{1-lim} often touches down with the least vertical velocity. The plots of $T_{sp}$ show the control policy responsible for this: the slow decay and few output spikes of \textit{1-lim} result in small `hops' that decrease in magnitude as the ground nears. Nevertheless, the landings performed by the single-spiking-neuron controller \textit{0-lim} also look promising. Like \textit{1-lim}, decoding decay is slower, which allows a larger number of acceleration setpoints to be selected. Still, the high frequency and large magnitude of the oscillations will most likely prevent a good transfer to the real world. Looking at \Cref{fig:realperf}, we see that this is indeed the case. The touchdown velocity of both \textit{20-base} and \textit{0-lim} is often higher than that of \textit{1-lim}, whose slower decoding dynamics helped with a successful transfer from simulation to the real world. Some unsteady behavior is still present during the final landing phase, however, as can be observed from the supplementary video. These `self-induced oscillations' are the result of the scale ambiguity in optical flow control~\cite{de_croon_monocular_2016}. Although SNN and ANN controllers are able to postpone these oscillations~\cite{scheper_evolution_2020}, getting rid of them completely requires additional measures~\cite{ho_adaptive_2018}.

\begin{figure}[!t]
    \centering
    \includegraphics[width=0.4\textwidth]{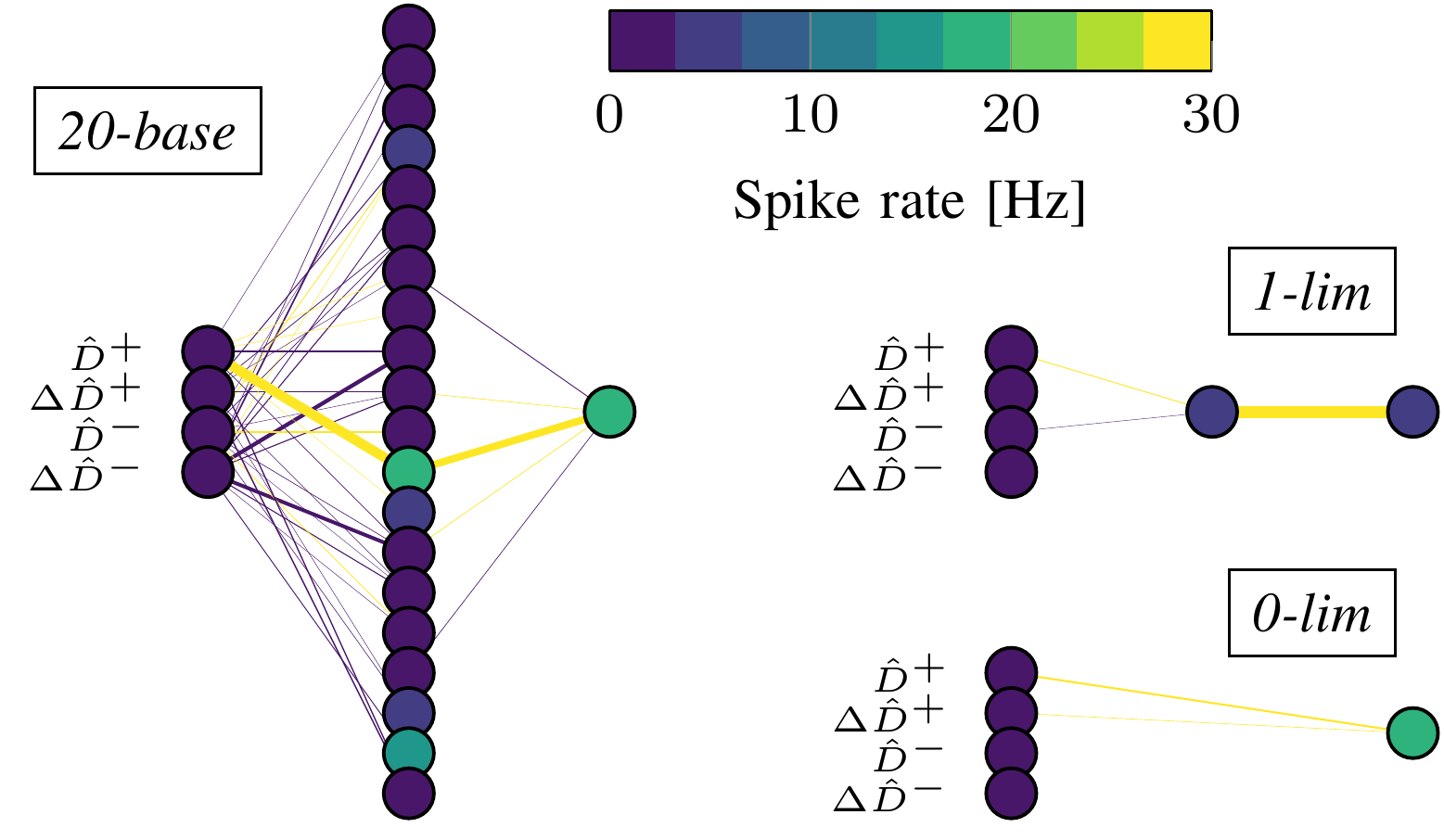}
    \caption{Average firing rates and synaptic weights of selected individuals for the five simulated runs displayed in \Cref{fig:simperf}. Vertex color is proportional to neuron firing rate, while synaptic weight is proportional to edge thickness. Edge colors indicate positive (yellow) or negative (purple) synapses.}
    \label{fig:networks}
\end{figure}

The network activity during simulated landings in \Cref{fig:networks} indicates that, in the case of \textit{0-lim}, further evolutionary optimization might decrease spike rate even more, as is shown to be feasible by the network of \textit{1-lim}. The same goes for \textit{20-base}, where the spiking of some hidden neurons is not used at all. This reflects in the large differences between the total network spike rates, which are 71.2~Hz, 7.5~Hz and 16.8~Hz for \textit{20-base}, \textit{1-lim} and \textit{0-lim}, respectively. Without spike minimization as an evolutionary objective, \textit{20-base} has a total spike rate of 201.2~Hz, meaning spike minimization is responsible for a 65\% drop. A further decrease is possible given more generations or a smaller network (an additional 31\% for \textit{1-lim}). Based on energy measurements for the neuromorphic Loihi chip~\cite{davies_loihi:_2018}, the corresponding energy savings would be 11.4~nJ (59\%) and 18.5~nJ (96\%) for spike minimization and smaller networks, respectively.

\Cref{fig:sstrans} compares the transient and steady-state response of the selected individuals. Due to its slow decoding dynamics, the transient response of \textit{1-lim} shows a much larger number of possible thrust setpoints than any other individual. Furthermore, it limits itself to a smaller $T_{sp}$ range, preventing large-magnitude oscillations. Both \textit{20-base} and \textit{0-lim}, on the other hand, only have a distinct number of plateaus in their transient response, and these have to cover the entire range $[-0.8, 0.5]$~g. Looking at the steady-state response, we see that \textit{20-base} and \textit{1-lim} mainly have a gradient in the $\hat{D}$-dimension, which makes sense given the connections in those networks to the respective encoding neuron. The fact that this gradient is mostly on the $+\hat{D}$-side suggests that an indication of positive divergence alone might suffice (absence of $+\hat{D}$ activity relates to $-\hat{D}$). The response of \textit{0-lim}, however, also has a significant gradient in the $\Delta\hat{D}$-dimension, as this individual additionally has a positive connection to the $\Delta\hat{D}^+$ input neuron.

The transient responses of the proposed controllers in \Cref{fig:sstrans} all seem to approximate a sigmoid shape. In comparison, the response given by a proportional divergence controller, whose output thrust is directly related to the divergence error, would be a straight line, with its slope dependent on the controller's gain. Analogously, the accompanying steady-state plot would show an even gradient along $\hat{D}$. The dotted black lines in \Cref{fig:sstrans} show P-controllers comparable to \textit{1-lim} and \textit{0-lim}.

\subsection{Comparison with Existing Controllers}

\begin{figure}[!t]
    \centering
    \includegraphics[width=0.49\textwidth]{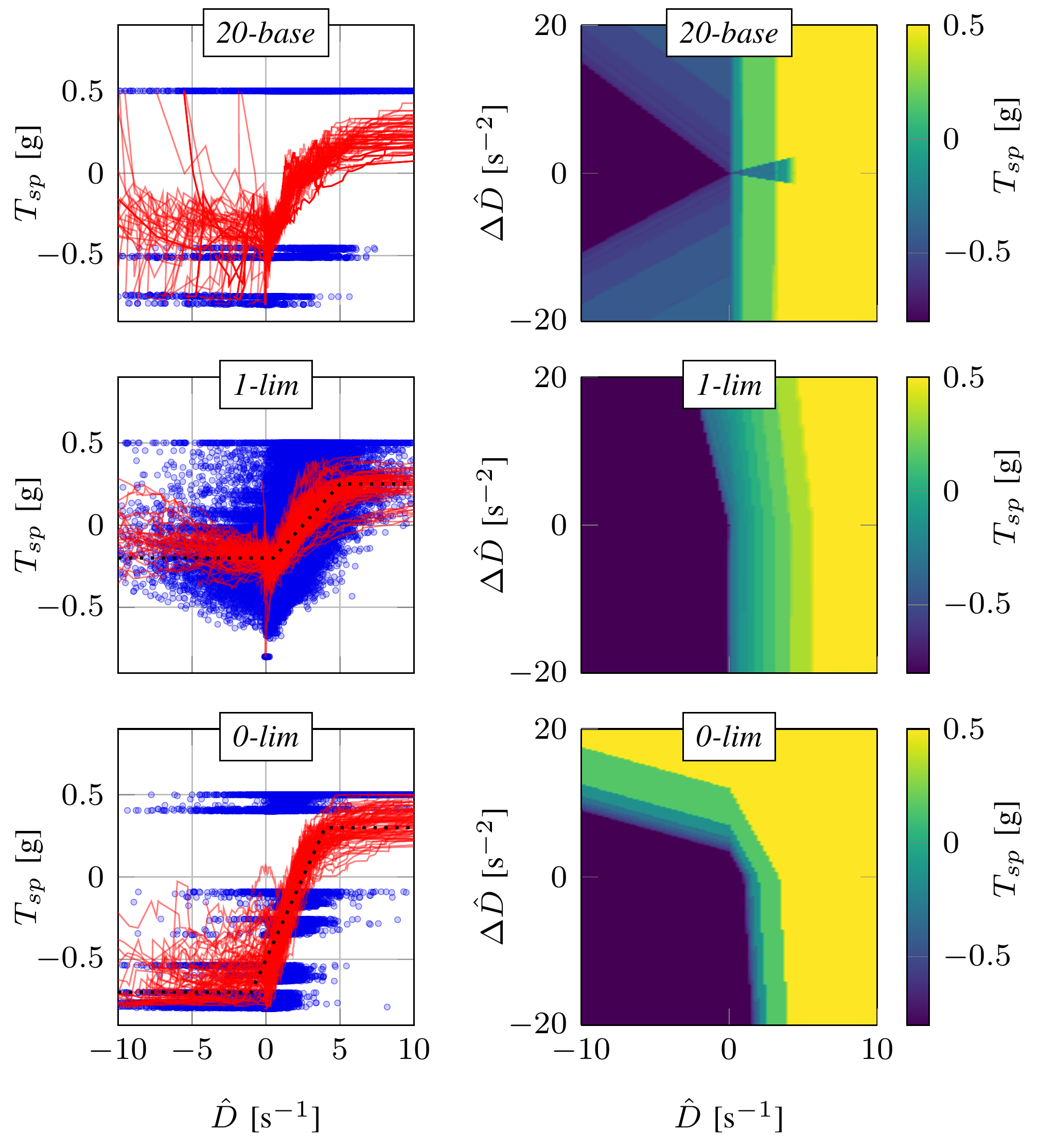}
    \caption{Transient and steady-state response of selected individuals. Steady-state responses are obtained by subjecting the SNNs to 100~time steps of the same observation and subsequently averaging the last 50~steps. The transient response is made up of 100~simulated landings during which $\hat{D}$ and $T_{sp}$ are recorded (blue dots), and then sorted by increasing divergence and passed through a 40-step moving average (red lines). Comparable proportional controllers are indicated by a dotted black line.}
    \label{fig:sstrans}
\end{figure}

\begin{figure}[!t]
    \hspace{0.4cm}
    \includegraphics[width=0.39\textwidth]{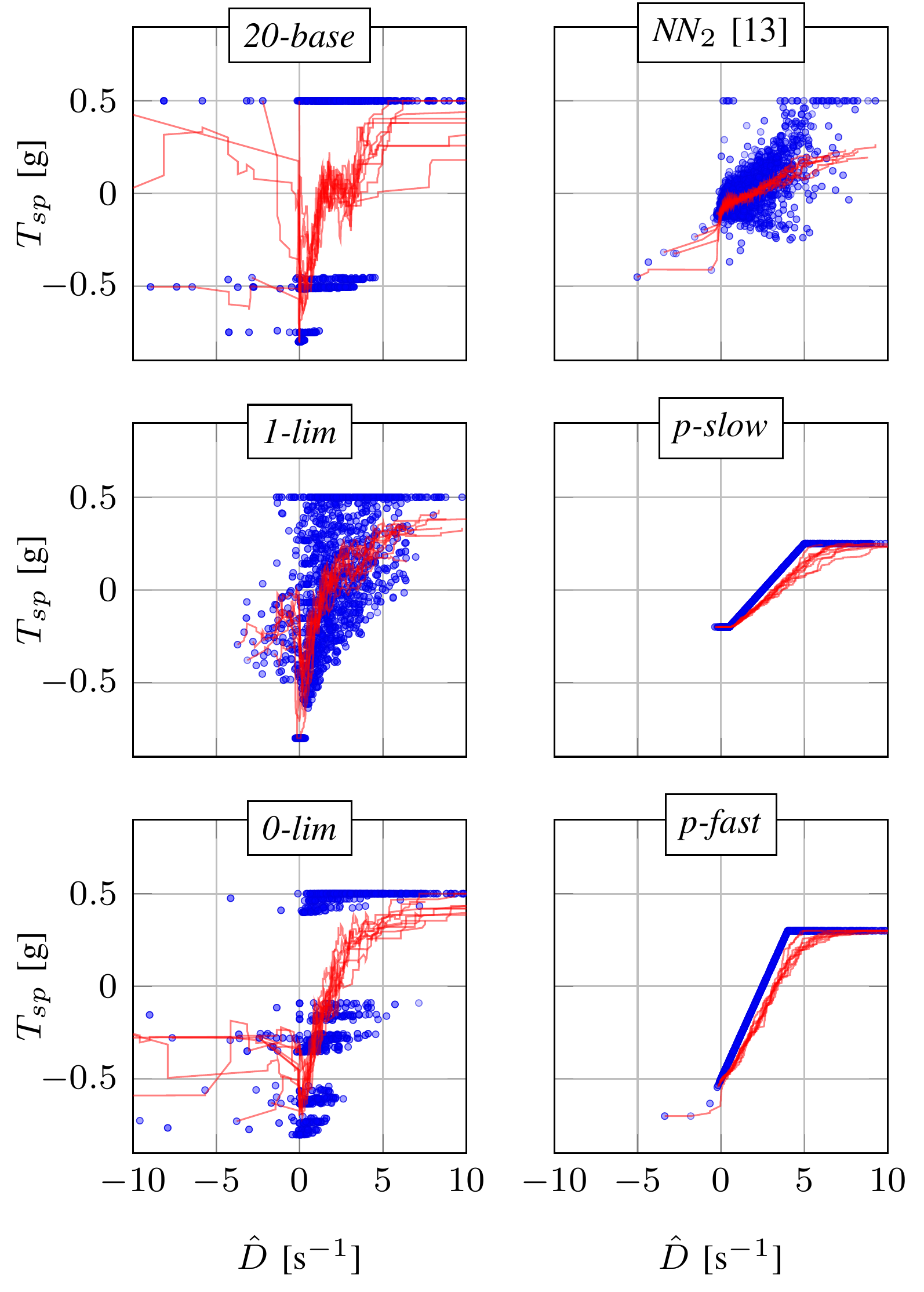}
    \caption{Comparison of transient responses from real-world tests. Responses are made up of ten real-world landings during which $\hat{D}$ and $T_{sp}$ are recorded (blue dots), and then sorted by increasing divergence and passed through a 40-step moving average (red lines).}
    \label{fig:nnpidtrans}
\end{figure}

To compare the performance of the proposed SNN controllers with existing control methods for divergence-based landing, we evaluate the transient responses and landing characteristics obtained from real-world flight tests. The right column of \Cref{fig:nnpidtrans} consists of current state-of-the-art control methods for optical flow landings. At the top, there is the \textit{NN$_{2}$} controller from \cite{scheper_evolution_2020}, an ANN evolved for divergence-based landing control with eight hidden neurons. The middle response is obtained from a pure P-controller, named \textit{p-slow}. Mathematically, the thrust output of a P-controller can be represented as $T_{sp} = \frac{K_p}{g} \cdot (\hat{D} - D_{sp})$. \textit{p-slow} has a gain $K_p = 0.98$ and a divergence setpoint $D_{sp} = 2.5$~s$^{-1}$, and its output $T_{sp}$ is clamped to a range $[-0.2, 0.25]$~g. Another P-controller, \textit{p-fast}, is included at the bottom, which instead has a gain $K_p = 1.96$ and a thrust clamping to $[-0.7, 0.3]$~g. Both \textit{p-slow} and \textit{p-fast} were derived from the transient responses of SNN controllers in terms of gain (slope) and divergence setpoint (offset), with the former being based on \textit{1-lim}, and the latter on \textit{0-lim} (see \Cref{fig:sstrans}).

Comparing the evolved SNN controllers with \textit{NN$_2$}, we see that the latter is characterized by a lower gain and a limited but high-resolution range of thrust setpoints, leading to slower but smooth landings. Looking at the P-controllers, the lack of stochasticity in their response is immediately obvious, making their landings smooth as well. Nonetheless, both \textit{1-lim} and \textit{0-lim} outperform their derived P-controllers \textit{p-slow} and \textit{p-fast} in terms of touchdown velocity while landing almost as quickly: 2.9~s / 0.4~ms$^{-1}$ versus 2.4~s / 1.0~ms$^{-1}$ and 2.2~s / 1.0~ms$^{-1}$ versus 1.9~s / 1.2~ms$^{-1}$, respectively. Also, \textit{1-lim} performed better than \textit{NN$_2$}, whose landings averaged 4.4~s / 0.5~ms$^{-1}$. Real-world landing profiles of \textit{1-lim}, \textit{NN$_2$} and \textit{p-slow} are shown in \Cref{fig:opticalflow}.

\section{Conclusion}
\label{sec:conc}

In this letter, we demonstrated, for the first time, that neuromorphic controllers evolved in a highly abstracted simulation environment are capable of controlling landings of real-world MAVs using only the divergence of the optical flow field. Further, by minimizing the amount of spikes during evolution, we provided insight into the resources required for successfully solving the problem at hand, and the potential energy savings of an implementation on neuromorphic hardware. A real-world comparison with state-of-the-art controllers showed that the proposed SNNs often land faster and touch down softer. Also, we found that SNNs consisting of only a single spiking neuron are equally capable of smooth landings as larger networks, all the while using only a fraction of their spikes. This is in line with~\cite{gidon_dendritic_2020}, which implies that single biological neurons are capable of solving linearly non-separable problems. Future research should focus on achieving an end-to-end spiking solution to vision-based control, making use of an SNN capable of estimating global motion from an event camera~\cite{paredes-valles_unsupervised_2019}.

\bibliographystyle{IEEEtran}
\bibliography{IEEEabrv,references}

\begin{thebibliography}{10}
\providecommand{\url}[1]{#1}
\csname url@samestyle\endcsname
\providecommand{\newblock}{\relax}
\providecommand{\bibinfo}[2]{#2}
\providecommand{\BIBentrySTDinterwordspacing}{\spaceskip=0pt\relax}
\providecommand{\BIBentryALTinterwordstretchfactor}{4}
\providecommand{\BIBentryALTinterwordspacing}{\spaceskip=\fontdimen2\font plus
\BIBentryALTinterwordstretchfactor\fontdimen3\font minus
  \fontdimen4\font\relax}
\providecommand{\BIBforeignlanguage}[2]{{%
\expandafter\ifx\csname l@#1\endcsname\relax
\typeout{** WARNING: IEEEtran.bst: No hyphenation pattern has been}%
\typeout{** loaded for the language `#1'. Using the pattern for}%
\typeout{** the default language instead.}%
\else
\language=\csname l@#1\endcsname
\fi
#2}}
\providecommand{\BIBdecl}{\relax}
\BIBdecl

\bibitem{gibson_perception_1950}
J.~J. Gibson, \emph{The {Perception} of the {Visual} {World}}.\hskip 1em plus
  0.5em minus 0.4em\relax Boston: Houghton Mifflin Company, 1950.

\bibitem{baird_universal_2013}
E.~Baird, N.~Boeddeker, M.~R. Ibbotson, and M.~V. Srinivasan,
  ``\BIBforeignlanguage{en}{A universal strategy for visually guided
  landing},'' \emph{\BIBforeignlanguage{en}{Proceedings of the National Academy
  of Sciences}}, vol. 110, no.~46, pp. 18\,686--18\,691, Nov. 2013.

\bibitem{posch_retinomorphic_2014}
C.~Posch, T.~Serrano-Gotarredona, B.~Linares-Barranco, and T.~Delbruck,
  ``Retinomorphic {Event}-{Based} {Vision} {Sensors}: {Bioinspired} {Cameras}
  {With} {Spiking} {Output},'' \emph{Proceedings of the IEEE}, vol. 102,
  no.~10, pp. 1470--1484, Oct. 2014.

\bibitem{de_croon_design_2009}
G.~C. H.~E. de~Croon, K.~M.~E. de~Clercq, R.~Ruijsink, B.~Remes, and
  C.~de~Wagter, ``\BIBforeignlanguage{en}{Design, {Aerodynamics}, and
  {Vision}-{Based} {Control} of the {DelFly}},''
  \emph{\BIBforeignlanguage{en}{International Journal of Micro Air Vehicles}},
  vol.~1, no.~2, pp. 71--97, Jun. 2009.

\bibitem{ma_controlled_2013}
K.~Y. Ma, P.~Chirarattananon, S.~B. Fuller, and R.~J. Wood,
  ``\BIBforeignlanguage{en}{Controlled {Flight} of a {Biologically} {Inspired},
  {Insect}-{Scale} {Robot}},'' \emph{\BIBforeignlanguage{en}{Science}}, vol.
  340, no. 6132, pp. 603--607, May 2013.

\bibitem{gallego_event-based_2019}
G.~Gallego, T.~Delbruck, G.~Orchard, C.~Bartolozzi, B.~Taba, A.~Censi,
  S.~Leutenegger, A.~Davison, J.~Conradt, K.~Daniilidis, and D.~Scaramuzza,
  ``Event-based {Vision}: {A} {Survey},'' \emph{arXiv:1904.08405 [cs]}, Apr.
  2019.

\bibitem{maass_networks_1997}
W.~Maass, ``Networks of spiking neurons: {The} third generation of neural
  network models,'' \emph{Neural Networks}, vol.~10, no.~9, pp. 1659--1671,
  Dec. 1997.

\bibitem{tavanaei_deep_2019}
A.~Tavanaei, M.~Ghodrati, S.~R. Kheradpisheh, T.~Masquelier, and A.~Maida,
  ``Deep learning in spiking neural networks,'' \emph{Neural Networks}, vol.
  111, pp. 47--63, Mar. 2019.

\bibitem{fremaux_neuromodulated_2016}
N.~Frémaux and W.~Gerstner, ``\BIBforeignlanguage{English}{Neuromodulated
  {Spike}-{Timing}-{Dependent} {Plasticity}, and {Theory} of {Three}-{Factor}
  {Learning} {Rules}},'' \emph{\BIBforeignlanguage{English}{Frontiers in Neural
  Circuits}}, vol.~9, 2016.

\bibitem{clawson_spiking_2016}
T.~S. Clawson, S.~Ferrari, S.~B. Fuller, and R.~J. Wood, ``Spiking neural
  network ({SNN}) control of a flapping insect-scale robot,'' in \emph{2016
  {IEEE} 55th {Conference} on {Decision} and {Control} ({CDC})}, Dec. 2016, pp.
  3381--3388.

\bibitem{bing_indirect_2020}
Z.~Bing, C.~Meschede, G.~Chen, A.~Knoll, and K.~Huang,
  ``\BIBforeignlanguage{en}{Indirect and direct training of spiking neural
  networks for end-to-end control of a lane-keeping vehicle},''
  \emph{\BIBforeignlanguage{en}{Neural Networks}}, vol. 121, pp. 21--36, Jan.
  2020.

\bibitem{zhao_brain-inspired_2018}
F.~Zhao, Y.~Zeng, and B.~Xu, ``\BIBforeignlanguage{English}{A
  {Brain}-{Inspired} {Decision}-{Making} {Spiking} {Neural} {Network} and {Its}
  {Application} in {Unmanned} {Aerial} {Vehicle}},''
  \emph{\BIBforeignlanguage{English}{Frontiers in Neurorobotics}}, vol.~12,
  2018.

\bibitem{scheper_evolution_2020}
K.~Y.~W. Scheper and G.~C. H.~E. de~Croon, ``\BIBforeignlanguage{en}{Evolution
  of robust high speed optical-flow-based landing for autonomous {MAVs}},''
  \emph{\BIBforeignlanguage{en}{Robotics and Autonomous Systems}}, vol. 124, p.
  103380, Feb. 2020.

\bibitem{floreano_neuroevolution:_2008}
D.~Floreano, P.~Dürr, and C.~Mattiussi,
  ``\BIBforeignlanguage{en}{Neuroevolution: from architectures to learning},''
  \emph{\BIBforeignlanguage{en}{Evolutionary Intelligence}}, vol.~1, no.~1, pp.
  47--62, Mar. 2008.

\bibitem{fogel_advantages_1997}
D.~B. Fogel, ``The {Advantages} of {Evolutionary} {Computation},'' in
  \emph{Biocomputing and {Emergent} {Computation}: {Proceedings} of
  {BCEC97}}.\hskip 1em plus 0.5em minus 0.4em\relax World Scientific Press,
  1997, pp. 1--11.

\bibitem{iglesias_dynamics_2005}
J.~Iglesias, J.~Eriksson, F.~Grize, M.~Tomassini, and A.~E.~P. Villa,
  ``\BIBforeignlanguage{en}{Dynamics of pruning in simulated large-scale
  spiking neural networks},'' \emph{\BIBforeignlanguage{en}{Biosystems}},
  vol.~79, no.~1, pp. 11--20, Jan. 2005.

\bibitem{dora_two_2015}
S.~Dora, S.~Sundaram, and N.~Sundararajan, ``A two stage learning algorithm for
  a {Growing}-{Pruning} {Spiking} {Neural} {Network} for pattern classification
  problems,'' in \emph{2015 {International} {Joint} {Conference} on {Neural}
  {Networks} ({IJCNN})}, Jul. 2015, pp. 1--7.

\bibitem{hwangbo_control_2017}
J.~Hwangbo, I.~Sa, R.~Siegwart, and M.~Hutter, ``Control of a {Quadrotor}
  {With} {Reinforcement} {Learning},'' \emph{IEEE Robotics and Automation
  Letters}, vol.~2, no.~4, pp. 2096--2103, Oct. 2017.

\bibitem{pfeiffer_deep_2018}
M.~Pfeiffer and T.~Pfeil, ``Deep {Learning} {With} {Spiking} {Neurons}:
  {Opportunities} and {Challenges},'' \emph{Frontiers in Neuroscience},
  vol.~12, Oct. 2018.

\bibitem{stanley_designing_2019}
K.~O. Stanley, J.~Clune, J.~Lehman, and R.~Miikkulainen,
  ``\BIBforeignlanguage{en}{Designing neural networks through
  neuroevolution},'' \emph{\BIBforeignlanguage{en}{Nature Machine
  Intelligence}}, vol.~1, no.~1, pp. 24--35, Jan. 2019.

\bibitem{howard_evolving_2014}
D.~Howard and A.~Elfes, ``\BIBforeignlanguage{en}{Evolving {Spiking} {Networks}
  for {Turbulence}-{Tolerant} {Quadrotor} {Control}},'' in
  \emph{\BIBforeignlanguage{en}{Artificial {Life} 14: {Proceedings} of the
  {Fourteenth} {International} {Conference} on the {Synthesis} and {Simulation}
  of {Living} {Systems}}}.\hskip 1em plus 0.5em minus 0.4em\relax The MIT
  Press, Jul. 2014, pp. 431--438.

\bibitem{howard_towards_2016}
D.~Howard and F.~Kendoul, ``\BIBforeignlanguage{en}{Towards {Evolved} {Time} to
  {Contact} {Neurocontrollers} for {Quadcopters}},'' in
  \emph{\BIBforeignlanguage{en}{Artificial {Life} and {Computational}
  {Intelligence}}}, ser. Lecture {Notes} in {Computer} {Science}.\hskip 1em
  plus 0.5em minus 0.4em\relax Cham: Springer International Publishing, 2016,
  pp. 336--347.

\bibitem{bing_survey_2018}
Z.~Bing, C.~Meschede, F.~Röhrbein, K.~Huang, and A.~C. Knoll,
  ``\BIBforeignlanguage{English}{A {Survey} of {Robotics} {Control} {Based} on
  {Learning}-{Inspired} {Spiking} {Neural} {Networks}},''
  \emph{\BIBforeignlanguage{English}{Frontiers in Neurorobotics}}, vol.~12,
  2018.

\bibitem{de_croon_optic-flow_2013}
G.~C. H.~E. de~Croon, H.~W. Ho, C.~De~Wagter, E.~van Kampen, B.~Remes, and
  Q.~P. Chu, ``\BIBforeignlanguage{en}{Optic-{Flow} {Based} {Slope}
  {Estimation} for {Autonomous} {Landing}},''
  \emph{\BIBforeignlanguage{en}{International Journal of Micro Air Vehicles}},
  vol.~5, no.~4, pp. 287--297, Dec. 2013.

\bibitem{ho_adaptive_2018}
H.~W. Ho, G.~C. H.~E. de~Croon, E.~van Kampen, Q.~P. Chu, and M.~Mulder,
  ``Adaptive {Gain} {Control} {Strategy} for {Constant} {Optical} {Flow}
  {Divergence} {Landing},'' \emph{IEEE Transactions on Robotics}, vol.~34,
  no.~2, pp. 508--516, Apr. 2018.

\bibitem{stein_theoretical_1965}
R.~B. Stein, ``A {Theoretical} {Analysis} of {Neuronal} {Variability},''
  \emph{Biophysical Journal}, vol.~5, no.~2, pp. 173--194, Mar. 1965.

\bibitem{liu_spike-frequency_2001}
Y.-H. Liu and X.-J. Wang, ``\BIBforeignlanguage{en}{Spike-{Frequency}
  {Adaptation} of a {Generalized} {Leaky} {Integrate}-and-{Fire} {Model}
  {Neuron}},'' \emph{\BIBforeignlanguage{en}{Journal of Computational
  Neuroscience}}, vol.~10, no.~1, pp. 25--45, Jan. 2001.

\bibitem{yao_evolving_1999}
X.~Yao, ``Evolving artificial neural networks,'' \emph{Proceedings of the
  IEEE}, vol.~87, no.~9, pp. 1423--1447, Sep. 1999.

\bibitem{deb_fast_2002}
K.~Deb, A.~Pratap, S.~Agarwal, and T.~Meyarivan, ``A fast and elitist
  multiobjective genetic algorithm: {NSGA}-{II},'' \emph{IEEE Transactions on
  Evolutionary Computation}, vol.~6, no.~2, pp. 182--197, Apr. 2002.

\bibitem{ho_characterization_2016}
H.~W. Ho and G.~C. H.~E. de~Croon, ``\BIBforeignlanguage{en}{Characterization
  of {Flow} {Field} {Divergence} for {MAVs} {Vertical} {Control} {Landing}},''
  in \emph{\BIBforeignlanguage{en}{{AIAA} {Guidance}, {Navigation}, and
  {Control} {Conference}}}.\hskip 1em plus 0.5em minus 0.4em\relax San Diego,
  California, USA: American Institute of Aeronautics and Astronautics, Jan.
  2016.

\bibitem{fortin_deap:_2012}
F.-A. Fortin, F.-M.~D. Rainville, M.-A. Gardner, M.~Parizeau, and C.~Gagné,
  ``{DEAP}: {Evolutionary} {Algorithms} {Made} {Easy},'' \emph{Journal of
  Machine Learning Research}, vol.~13, no. Jul, pp. 2171--2175, 2012.

\bibitem{pijnacker_hordijk_vertical_2018}
B.~J. Pijnacker~Hordijk, K.~Y.~W. Scheper, and G.~C. H.~E. de~Croon,
  ``\BIBforeignlanguage{en}{Vertical landing for micro air vehicles using
  event-based optical flow},'' \emph{\BIBforeignlanguage{en}{Journal of Field
  Robotics}}, vol.~35, no.~1, pp. 69--90, 2018.

\bibitem{rosten_machine_2006}
E.~Rosten and T.~Drummond, ``\BIBforeignlanguage{en}{Machine {Learning} for
  {High}-{Speed} {Corner} {Detection}},'' in
  \emph{\BIBforeignlanguage{en}{Computer {Vision} – {ECCV} 2006}}, ser.
  Lecture {Notes} in {Computer} {Science}.\hskip 1em plus 0.5em minus
  0.4em\relax Springer Berlin Heidelberg, 2006, pp. 430--443.

\bibitem{bouguet_pyramidal_2000}
J.-Y. Bouguet, ``Pyramidal {Implementation} of the {Affine} {Lucas} {Kanade}
  {Feature} {Tracker},'' 2000.

\bibitem{de_croon_monocular_2016}
G.~C. H.~E. de~Croon, ``\BIBforeignlanguage{en}{Monocular distance estimation
  with optical flow maneuvers and efference copies: a stability-based
  strategy},'' \emph{\BIBforeignlanguage{en}{Bioinspiration \& Biomimetics}},
  vol.~11, no.~1, p. 016004, Jan. 2016.

\bibitem{davies_loihi:_2018}
M.~Davies, N.~Srinivasa, T.~Lin, G.~Chinya, Y.~Cao, S.~H. Choday, G.~Dimou,
  P.~Joshi, N.~Imam, S.~Jain, Y.~Liao, C.~Lin, A.~Lines, R.~Liu,
  D.~Mathaikutty, S.~McCoy, A.~Paul, J.~Tse, G.~Venkataramanan, Y.~Weng,
  A.~Wild, Y.~Yang, and H.~Wang, ``Loihi: {A} {Neuromorphic} {Manycore}
  {Processor} with {On}-{Chip} {Learning},'' \emph{IEEE Micro}, vol.~38, no.~1,
  pp. 82--99, Jan. 2018.

\bibitem{gidon_dendritic_2020}
A.~Gidon, T.~A. Zolnik, P.~Fidzinski, F.~Bolduan, A.~Papoutsi, P.~Poirazi,
  M.~Holtkamp, I.~Vida, and M.~E. Larkum, ``\BIBforeignlanguage{en}{Dendritic
  action potentials and computation in human layer 2/3 cortical neurons},''
  \emph{\BIBforeignlanguage{en}{Science}}, vol. 367, no. 6473, pp. 83--87, Jan.
  2020.

\bibitem{paredes-valles_unsupervised_2019}
F.~Paredes-Vallés, K.~Y.~W. Scheper, and G.~C. H.~E. de~Croon, ``Unsupervised
  {Learning} of a {Hierarchical} {Spiking} {Neural} {Network} for {Optical}
  {Flow} {Estimation}: {From} {Events} to {Global} {Motion} {Perception},''
  \emph{IEEE Transactions on Pattern Analysis and Machine Intelligence}, 2019.

\end{thebibliography}

\end{document}